\documentclass{acm_proc_article-sp}

\DeclareMathOperator*{\argmin}{\arg\!\min}

\usepackage{algorithm}
\usepackage{algpseudocode}
\begin{document}

\title{Classification of Passes in Football Matches using Spatiotemporal Data\titlenote{
        A poster submission has also been made to the Large Scale Sports Analytics workshop at the KDD conference on this work. This submission covers the problem definition and solution framework described in this paper, and also includes preliminary experimental results on the accuracy of the framework.
    }
}

%
%
%
%

\numberofauthors{4} 
%
\author{
\alignauthor 
Michael Horton\\
       \affaddr{mhor9676@uni.sydney.edu.au}\\
\alignauthor
Joachim Gudmundsson\\
       \affaddr{joachim.gudmundsson@\\sydney.edu.au}\\
\alignauthor
Sanjay Chawla\\
       \affaddr{sanjay.chawla@sydney.edu.au}\\
       \and
\alignauthor 
Jo\"{e}l Estephan\\
    \affaddr{j.estephan4@gmail.com}\\
%
       \affaddr{School of Information Technologies}\\
       \affaddr{The University of Sydney}\\
       \affaddr{Sydney, Australia}\\
}

\maketitle

\begin{abstract}
\label{sec:abstract}

A knowledgeable observer of a game of football (soccer) can make a subjective evaluation of the quality of passes made between players during the game. We investigate the problem of producing an automated system to make the same evaluation of passes. We present a model that constructs numerical predictor variables from spatiotemporal match data using feature functions based on methods from computational geometry, and then learns a classification function from labelled examples of the predictor variables. Furthermore, the learned classifiers are analysed to determine if there is a relationship between the complexity of the algorithm that computed the predictor variable and the importance of the variable to the classifier. Experimental results show that we are able to produce a classifier with $85.8\%$ accuracy on classifying passes as \emph{Good}, \emph{OK} or \emph{Bad}, and that the predictor variables computed using complex methods from computational geometry are of moderate importance to the learned classifiers. Finally, we show that the inter-rater agreement on pass classification between the machine classifier and a human observer is of similar magnitude to the agreement between two observers.
\end{abstract}



 
\section{Introduction}
\label{sec:introduction}

%
There is considerable research into developing objective methods of analysing sports, including football. Football analysis has practical applications in player evaluation for coaching and scouting; development of game and campaign strategies; and also to enhance the viewing experience of televised matches. Currently, football analysis is typically done manually or using simple frequency analysis. There would thus appear to be scope to improve the efficiency of the analysis process as well as the quality of the output.

The contribution that we intend to make is to use methods from computational geometry to compute complex predictor variables and to then apply machine learning classification algorithms to produce quantitative measures of passing performance. This combination is relatively novel and if successful, may be applied to domains other than football.

The remainder of the paper is structured as follows. Section~\ref{sec:problem-definition} defines the problem and the model we propose to solve it. The feature functions that compute the predictor variables used for pass classification are described in Section~\ref{sec:feature-engineering} and the process for obtaining labelled examples in Section~\ref{sec:data}. The learning algorithms used are described in Section~\ref{sec:classification}. The experimental setup and results are reported in Section~\ref{sec:experiments} and an analysis of the results and areas of future work is provided in Section~\ref{sec:analysis}.

\subsection{Related Work}

Sports analysis using computational and probabilistic processes is a relatively recent innovation \cite{reilly2003}. The initial focus was on sports with more structured play, such as baseball and cricket, where collection of raw performance data is a more straightforward proposition. Within the past decade, advances in computer vision and image processing have resulted in systems that can capture accurate positional data about players in less structured games like football, hockey and basketball. However, the analytic capabilities that have been built on this type of data are currently fairly limited. 


One of the primary roles of a football coach is to analyse the performance of players during matches. However, experiments have found that the level of recollection of critical events in football matches is as low as 42\%~\cite{franks1986}. Given this level of recall, the use of a systematic approach to collecting data about events occurring during matches is desirable. 

The performance evaluation on these data have typically been through frequency analysis of events, such as the number of passes made by a player, or the number of unforced errors \cite{borrie2002}. However, the actions taken by players in a match are influenced by the interactions with other players, and simple frequency data may not fully capture the situation.

Event data represents only a small proportion of the data that could conceivably be collected from football matches. In order to gather sufficient data to perform complex analysis much research has focused on systems that capture positional data on the players and ball with high frequency and precision. The volumes of data inherent in such systems make it necessary to apply computerised solutions.

Collecting positional data is possible using modern image capture and processing technology~\cite{taki2000,leo2008}. There are commercial companies with similar systems such as Prozone \cite{prozone-2013} and Opta \cite{web:opta2013} which are used in professional leagues. State of the art systems are currently able to do this with high accuracy, for example the Prozone system is able to provide player co-ordinates accurate to 10cm at an frequency of 10Hz~\cite{prozone-2013}. 

The availability of accurate positional and event data provides the basis for complex analysis of matches and player performance. To date, a range of diverse approaches have been explored.

Pattern matching has been used in a number of applications. Borrie et al. uses T-pattern detection to extract similar passing sequences from matches~\cite{borrie2002}. Gudmundsson and Wolle analysed sub-trajectories using the Fr\'{e}chet distance as a metric to cluster sub-trajectories that occurred multiple times~\cite{gudmundsson2010}. Furthermore, Gudmundsson and Wolle also used suffix trees to encode sequences of passes between distinct players so that common sequences can be clustered~\cite{gudmundsson-wolle2012}.

Taki and Hasegawa defined a geometric subdivision called a \emph{dominant region}~\cite{taki2000}, akin to a Voronoi region~\cite{deberg2008}, that subdivides the pitch into cells owned by players such that the player can reach all areas of the cell before any other player. The dominant region considers the direction and velocity of the players~\cite{deberg2008}. This concept is further developed by Fujimura and Sugihara~\cite{fujimura2005} and Nakanishi et al.~\cite{nakanishi2010} who define efficient approximation models for calculating the dominant region. Gudmundsson and Wolle consider a related problem of computing the passing options that the player in possession has, based on the areas of the pitch that are reachable by the other players~\cite{gudmundsson-wolle2012}. Kang et al. uses a simplified motion model to compute a similar subdivision in order to define a set of performance measures for assessing player's passing and receiving performance~\cite{kang2006}. The defined measures are easily understandable and could thus be applied directly to rating player performance.

\section{Problem Definition}
\label{sec:problem-definition}
We consider the problem of classifying passes made during a football match based on the quality of the pass. The approach that we selected was to use supervised machine learning algorithms to learn a classification function. Such algorithms accept data of a particular structure, and thus a significant component of our approach is to take the raw spatiotemporal data and transform it to an appropriate format for the learning algorithms.

The input to the framework consists of three datasets for each match: a set of spatiotemporal trajectory sequences for each player; a sequence of events that occurred during the match; and a mapping of players to their respective teams. We used data from four matches played in the English Premiership during the 2007/08 season. The data was provided by Prozone \cite{prozone-2013}. These matches contained 2,932 passes in total.

The input data is used to construct a vector of predictor variables for each pass made during the match. In addition, the quality of the passes made in the four matches have been manually labelled by human observers. These labels are used to train the classifiers.

\subsection{Preliminaries}
We formally set up the problem by first defining the notation.  

We will focus on one match and assume that there is a global clock which increment in steps of size $s$ where $S = \{s_i | i \in \mathbb{N}\}$. For example, in our case, the data is updated at a frequency of 10Hz.

Let $P = \{p_1, p_2, ..., p_n\}$ be the set of players who appeared in the match. Associated with each player $p$ is a trajectory $t_{p} = \{(x_{i},y_{i}) |  i \in TI(p)\}$, where $TI(p)$ is the time interval when the player $p$ is on the pitch, i.e. has not been substituted. To access the geometry of the player $p$ at time step $s$, we will use the notation: $x(t_{p}, s)$ for the co-ordinates in $\mathbb{R}^2$; $\gamma(t_{p}, s)$ for the angle that the player is facing in where $\gamma \in [-\pi, \pi)$; and $v(t_{p}, s)$ for the velocity of the player. We denote the set of trajectories for all players in $P$ as $Tr$.

We associate a set of events $E$ with a game.  Each event $e \in E$ has a type (from a fixed-set), such as a pass, a shot, a tackle, etc. 
Formally, an event $e$ is a three-tuple $e = (s, u, R)$. Here $s$ is the time step when the event occurred, $u$ is the type of the event and $R$ is the set of players involved in the event. A set of events that are of particular interest are the pass events which are characterized as $\{(s,u,\{p\})|u = Pass, p \in P\}$, where $p$ is the player who makes the pass. Note, $s(e)$ is defined as the time when the event $e$ occurred.

Now, the predictor variables associated with each event are defined as follows. The predictor variables can (in principle) depend upon all the previous and future events. Similarly, a predictor variable can depend on the current location of all players, and possibly their previous and future locations.
%
Thus, the trajectory sequences, event sequence and player mappings are used to compute the predictor variables for each pass event. Let $\phi_j$ be the feature function that produces the $j$-th predictor variable: 
\begin{align*}
    \phi_j & : S \times Tr \times E  \times M \rightarrow \mathbb{R} 
\end{align*}
It is convenient to work with a vector of predictor variables for the $i$-th pass event:
\begin{align*}
    x^{(i)} & =  
    \begin{bmatrix}
        \phi_1(s(e_i), Tr, E, M) \\
        \phi_2(s(e_i), Tr, E, M) \\
        \vdots \\
        \phi_n(s(e_i), Tr, E, M)
    \end{bmatrix} \\
    x^{(i)}  & \in \mathbb{R}^{n}
\end{align*}
Finally, in order to train the learning algorithms, example labels are required for each class as response variables. The response variable is drawn from a discrete set of size $k$: $y \in Y$, and $|Y| = k$. Combining the response variable vectors with the corresponding predictor variables provides the training examples matrix for the machine learning algorithms.
\begin{align*}
    X  & = 
    \begin{bmatrix}
        \rule[.5ex]{2.5ex}{0.5pt} (x^{(1)})^{\top} \rule[.5ex]{2.5ex}{0.5pt} \\
        \rule[.5ex]{2.5ex}{0.5pt} (x^{(2)})^{\top} \rule[.5ex]{2.5ex}{0.5pt} \\
        \vdots \\
        \rule[.5ex]{2.5ex}{0.5pt} (x^{(m)})^{\top} \rule[.5ex]{2.5ex}{0.5pt} \\
    \end{bmatrix} &
    y  & = 
    \begin{bmatrix}
        y_1 \\
        y_2 \\
        \vdots \\
        y_m
    \end{bmatrix} \\
    X & \in \mathbb{R}^{m \times n} &
    y & \in \mathbb{R}^{n}
\end{align*}
Formally, we can define the classification problem as follows: Given the training set of pass events $X$ and labels $y$, and a cost function $J(X,y,\theta)$, the objective is to learn a parameterisation, $\theta$ such that:
\[
    \theta^{\ast} = \argmin_{\theta} J(X,y,\theta)
\]
This parameterisation characterizes the classifier function $h_{\theta}(x)$ that will predict the response variable $y$, given the input vector $x$.

\section{Predictor Variables}
\label{sec:feature-engineering}

The challenge of the predictor variable engineering task is to extract sufficient information from the spatiotemporal match data so that the classifiers are able to make accurate inferences about the quality of the passes made in a match. The objective is thus to ensure that sufficient information about the match state is inherent in the constructed predictor variables.

To place this problem in context, consider how an informed observer of a football match would make an assessment about the quality of a pass.

At a basic level, the observer would consider the fundamentals of the pass, such as the distance and speed of the pass and whether the intended recipient of the pass was able to control the ball. These are the basic geometric aspects of the pass, but even at this level, the observer is required to make some inferences, such as who the intended recipient of the pass is.

They would also likely consider the context of the match state when the pass was made. For example, was the passer under pressure from opposition players? To make such assessments, the observer would consider the positions of the players and the speed and direction that they were moving in, and the observer would make assumptions about whether the players are physically able to influence the pass by pressuring the passer or intercepting the pass. The observer thus has an idea of the physiological capabilities of the players, and will consider this in their estimation of the quality of a pass made.

At a higher level, players in a football match do not make passes simply to move the ball from one location to another. They make passes to improve the strategic position of their team. Passes can be made to improve the position of the ball, typically by trying to move the ball closer to the opponent's goal in order to have an opportunity to score. Passes may also be made to improve the match state by moving the ball from a congested area of the pitch to an area where the team in possession has a numerical or positional advantage. Meanwhile, the opposition will be actively trying to reduce the options of the player in possession to make passes. Thus, the match observer would need to consider the tactical and strategic objectives of the passer, and thus would have an understanding of the tactics and strategies employed by the player and team, and apply them to their estimate. Likewise, the observer would consider the defensive team and their strategies and tactics.

A football match can also be viewed as a sequence of events occurring at particular times. Examples of these events are touches, passes, tackles, fouls, shots on goal, and goals. The event-type that we are concerned with, \textit{Pass}, can thus be viewed as part of a sequence. This sequence can be subdivided in various ways, for example by unbroken sub-sequences of events where a single player or team is in control of the ball, or by a sub-sequence of events that occur between stoppages in play such as fouls, goals or injuries. When assessing the quality of a pass, the observer may consider the context of the pass in the sequence of events.

Finally, the observer may also consider the opportunity cost of the pass. By making the pass, the player forsook the other options available, such as passing to other players, dribbling or shooting.

For the trained observer, synthesising all this disparate information and making a prediction is a mental exercise that can be done in a matter of seconds. The problem described in this paper is to replicate this in a computational process. This section discusses the approach chosen to construct a set of predictor variables from the spatiotemporal and event data.

\subsection{Feature Functions}

Feature functions are used to compute the predictor variables that are input to the classifier. The feature function $\phi_j(s(e_i), Tr, E, M)$ outputs the $j$-th predictor variable for the $i$-th pass event. The predictor variables are divided into the following categories in a manner consistent with our analysis of the types of information used by the informed observer:

\begin{description}
  \item[Basic geometric predictor variables] are the predictor variables derived from the basic orientation of the players and ball on the football pitch. The feature functions for these predictor variables implement simple geometric operations such as determining angles between points, measuring Euclidean distances, and calculating velocity of objects over a time interval.
  \item[Sequential predictor variables] are constructed from the event sequence data. Currently three types of sequences are modelled: player possession, where a single player is in possession; team possession, where players from the same team are in uninterrupted possession; and play possession, where events between stoppages are grouped in a sequence. Examples of these predictor variables are the ordinal position of the event in a sequence, the duration of the sequence, and the event that is the final outcome of the current sequence.
  \item[Physiological predictor variables] are predictor variables that incorporate some aspect of the physiological capabilities of the players, generally how quickly they can reach a given point. Inherent in these predictor variables is a motion model that simulates the physical capabilities of the players. This is discussed in detail in Section~\ref{subsection:player-motion-model}.
  \item[Strategic predictor variables] are predictor variables that are designed to provide some information about the strategic element of the football match. The approach taken was to design predictor variables based on the dominant region construct proposed by \cite{taki2000}, see Section~\ref{subsection:dominant-region}.
\end{description}

The physiological and strategic predictor variables required further data structures for their computation, described in the following paragraphs.

\subsection{Player Motion Model}
\label{subsection:player-motion-model}

Consider a player at a certain point, and moving at a given velocity and direction. The time that it would take this player to reach another given point would intuitively appear to be of interest. In particular, if a player can reach a point before any other player, then that player dominates that point. This notion is the basis for the physiological and strategic predictor variables we have defined in our model. In order to determine the time required to reach a given point, a motion model of the player is required.

We examined the three motion models. They are intended to approximate the \textit{reachable region} of points that a player can reach in a given time. The time that it would take for a player positioned at a certain point to reach another point is determined by a number of factors. The maximum achievable velocity of the player would be a factor, as would the speed and direction the player is already heading in. Indirect factors could also play a part, such as the condition of the pitch, the wind direction and speed. These factors could all impact the reachable region, and we propose an approximation function for this.

Formally, the motion model can be defined as a function $f_{p}$ for a player $p$ that takes as arguments: the coordinates of a point in $\mathbb{R}^{2}$, expressed in polar form $(\theta, d)$ relative to the player's current location; the direction the player is facing $\gamma$; and the current velocity of the player $v$. The function returns the time $t$ it would take for the player to reach the point.
\begin{equation}
  \label{eq:continuous-motion-function}
  f_{p}(\theta, d, \gamma, v) = t
\end{equation}

\begin{figure*}
  \centering
  \begin{minipage}{0.3\textwidth}
    \includegraphics[width=0.8\linewidth, natwidth=300, natheight=300]{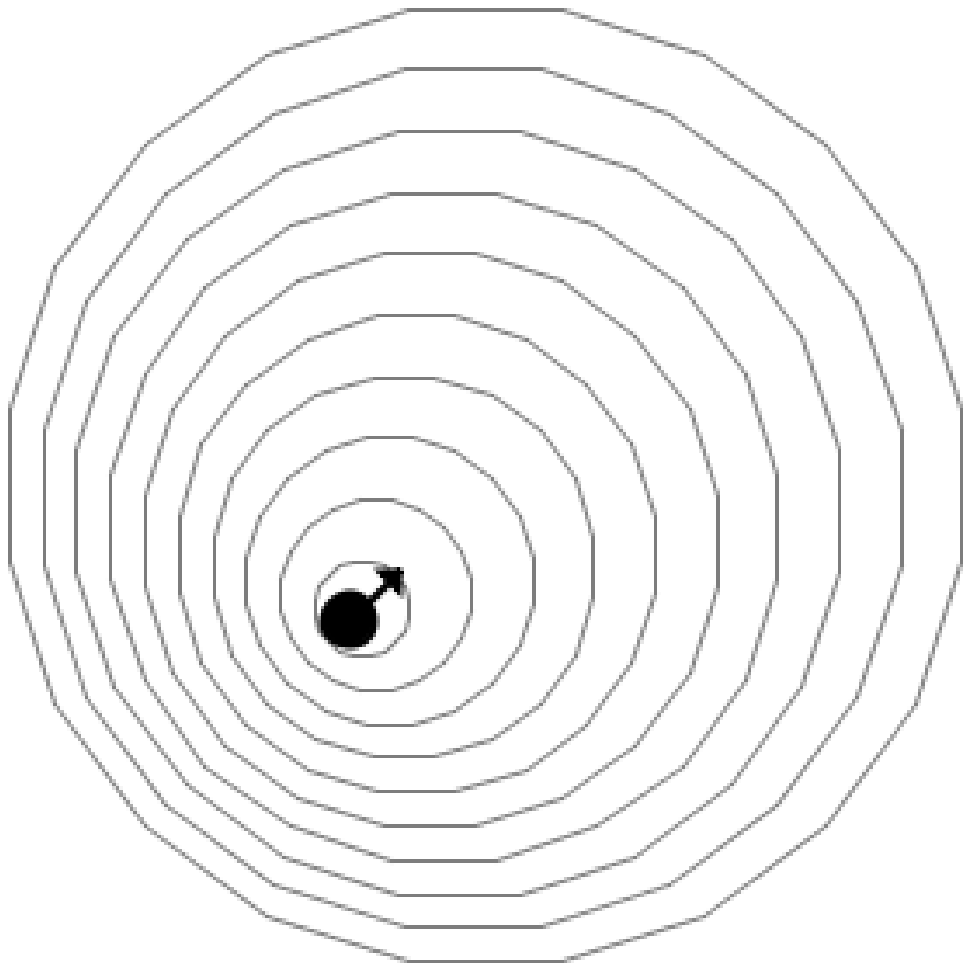}
    \label{figure:circle-motion-model}
  \end{minipage}
  \begin{minipage}{0.3\textwidth}
    \includegraphics[width=0.8\linewidth, natwidth=300, natheight=300]{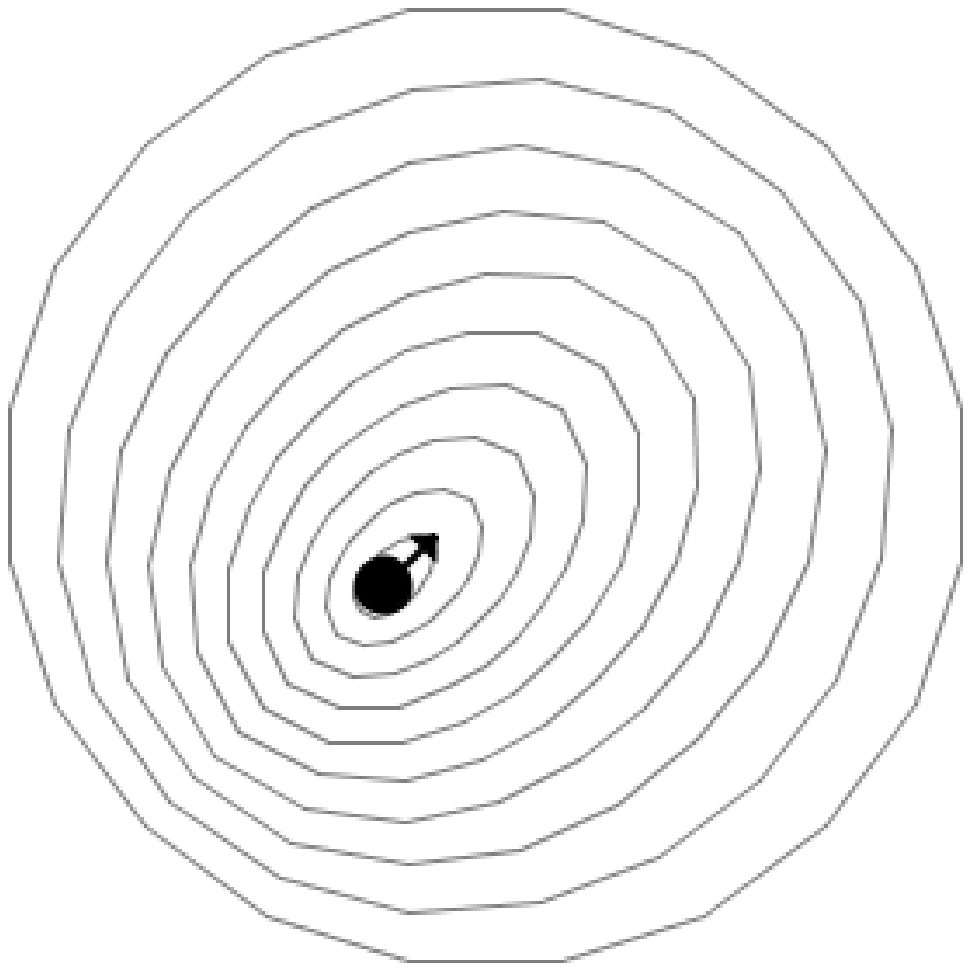}
    \label{figure:ellipse-motion-model}
  \end{minipage}
  \begin{minipage}{0.3\textwidth}
    \includegraphics[width=0.8\linewidth, natwidth=300, natheight=300]{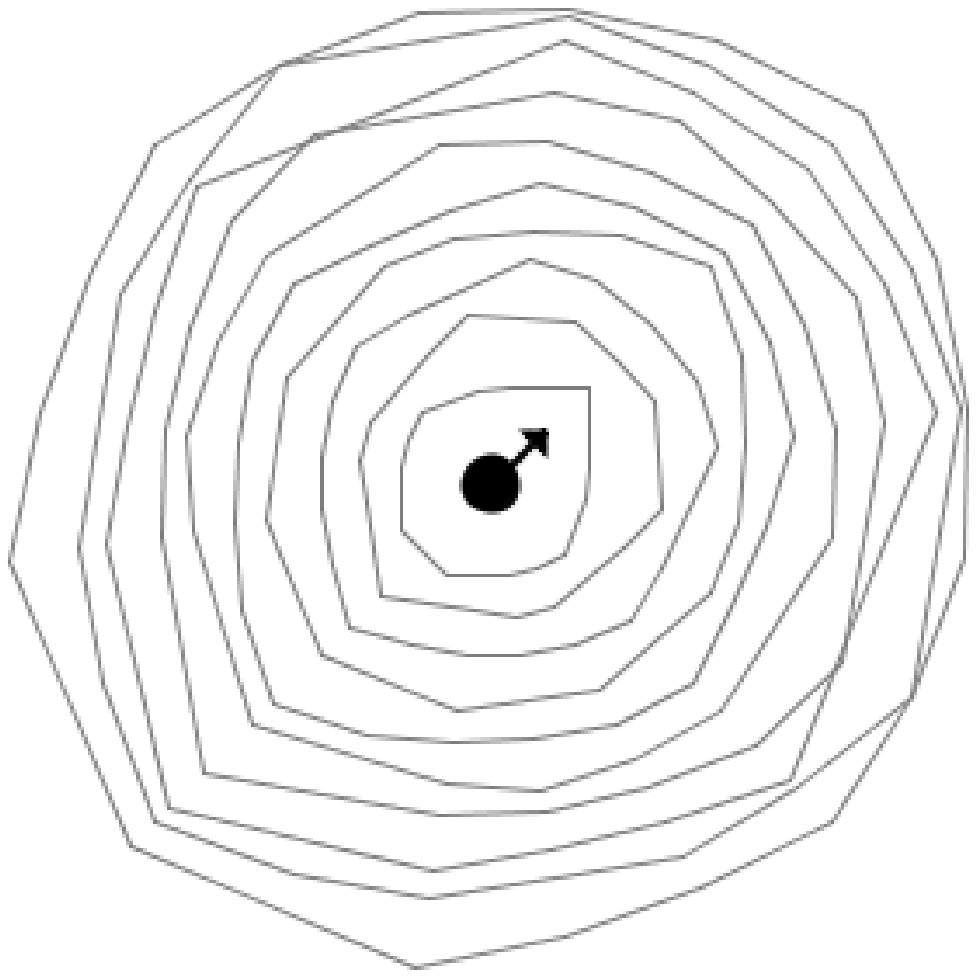}
    \label{figure:data-motion-model}
  \end{minipage}
  \caption{Reachable region boundaries using different motion models. (a) Circle, (b) Ellipse \& (c) Data-driven}
  \label{fig:player-reachable-region}
\end{figure*}
Gudmundsson and Wolle~\cite{gudmundsson-wolle2012} propose three simple motion models to approximate the reachable region for a player. The models discretize the reachable region by introducing a time-step $\tau \in \mathrm{T}$ to approximate $t$. The motion model would thus map a series of boundary curves that surround the initial starting point of the player. Furthermore, each curve is approximated by an $n$-sided polygon. The three motion models considered are based on a circular boundary, an elliptical boundary and a boundary constructed by sampling the from the trajectory sequences, see Figure~\ref{fig:player-reachable-region}. 

The motion models are used in several of the feature functions, in particular those based on the dominant region, below. The extended version of this paper evaluated these three motion models~\cite{horton-2013}, however the experiments here are based on the ellipse motion model, as this model produced classifiers whose performance in the experiments was generally close to being optimal. 

\subsection{The Dominant Region}
\label{subsection:dominant-region}

Taki and Hasegawa presents the dominant region as a dynamic area of influence that a player in a football match can exert dominance over, where dominance is defined as being the regions of the pitch that the player is able to reach before any other player~\cite{taki2000}. We propose to use the dominant region as a measure to approximate the strategic position that a team at a given point in time in the match by constructing feature functions based on it. Furthermore, we use the dominant region to also construct predictor variables that model the pressure exerted on the player in possession of the ball by opposition players in close proximity. 

Intuitively, these would appear to be useful predictors for the task of rating passes. The passing player wants to put the ball at a point where the intended recipient can reach it first, and this is, by definition, in the receiving player's dominant region. Thus, the proportion of the pitch that the team in possession dominates is a factor in the passing options of the passing player. Similarly, the size of the dominant region surrounding the passing and receiving players would provide information about the pressure the player is under.

The football pitch can be partitioned into an subdivision of dominant regions, each dominated by a particular player. It is thus conceptually similar to the Voronoi region \cite{deberg2008}, the difference being the function that determines the region that a particular point belongs to. The function for a Voronoi region is the Euclidean distance, and for the dominant region is the time it takes for a player to reach a given point. 

Formally, from \cite{taki2000}, a dominant region is defined by the following equation for a player $p_k$ at time-step $s$: 
\begin{multline}
  \label{eq:dominant-region}
  D(p_{k}, s) = \{x \in \mathbb{R}^2 | g(x, p_{k}, s) \leq g(x, p_{m}, s) \\
  \mbox{~for~} m \neq k, p_m \in P, s \in \mathbb{T}\}
\end{multline}

\begin{figure}
  \centering
    \includegraphics[width=1\linewidth, natwidth=1000, natheight=660]{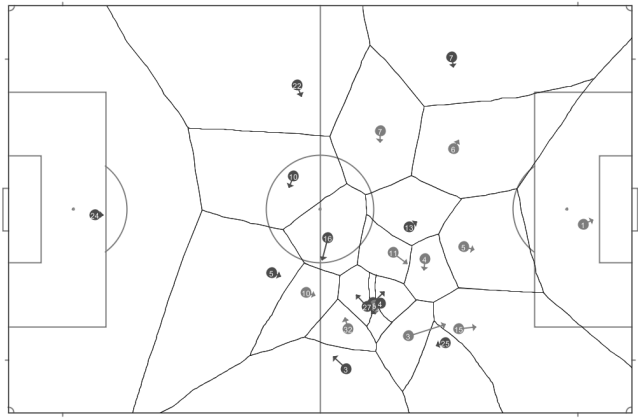}
  \caption{Dominant regions constructed using ellipse motion model.}
  \label{figure:dominant-region}
\end{figure}

In the described setting, Equation~\eqref{eq:continuous-motion-function} determines the time taken for a player to reach a point $(\theta, d)$. However, this function is independent of the actual location of player $p$, and points are represented in polar form. To reconcile this, we define the following: $\theta(x_1, x_2)$ accepts as arguments two points in Euclidean form and computes the angle $\theta$ between them; and, likewise, $d(x_1, x_2)$ computes the Euclidean distance between the points.
Thus, $g$ is defined as:
\begin{multline*}
    g(x, p_{k}, s) \mapsto f_{p_k}(\theta(x(t_{p_{k}}, s), x), d(x(t_{p_{k}}, s), x), \\
    \gamma(t_{p_{k}}, s), v(t_{p_{k}}, s)) 
\end{multline*}
%
The subdivision of the dominant regions for all players will thus partition the football pitch, as can be seen in Figure~\ref{figure:dominant-region}. However, $x \in \mathbb{R}^2$ is continuous, and there is currently no algorithm available to efficiently compute this continuous function. In theory, the dominant region of a player may not even be a single connected region~\cite{taki2000}. Computing the intersection of surfaces in three dimensions, as required in Equation~\eqref{eq:dominant-region} is non-trivial and time-consuming. As such, we use an approximation algorithm to compute the dominant region. Taki and Hasegawa~\cite{taki2000} and Nakanishi et al.~\cite{nakanishi2010} both present approximation algorithms where $x$ is approximated by a discrete grid $Y \subset \mathbb{N}^2$, and the dominant regions are thus computed for all points in $Y$. In the following subsection we present an approximation algorithm that uses the approximate motion model function to compute the boundaries of the dominant regions. 

\subsection{Approximation Algorithm}
\label{subsection:dominant-region-approx-algorithm}


In this section we present an approximation algorithm for efficiently computing the dominant region arrangement of the football pitch at a given point of time.  The algorithm, presented in Algorithm~\ref{alg:dominant-region}, has three steps. First, for every pairwise combination of players, the intersection points between the reachable region polygons are determined. In the second step, the intersection points are used to produce a reachable boundary between each pair of players. The third step constructs the smallest enclosing polygon around each player from the boundaries, and this is the player's dominant region.

\begin{algorithm}
    \caption{Approximation algorithm to compute the dominant region at a given time.}
    \label{alg:dominant-region}
    \begin{algorithmic}
        \State $P \gets \{p_o, \cdots, p_n\}$
        \Comment{The players}
        \State $T \gets \{\tau_0, \cdots, \tau_{max}\}$
        \Comment{The time-steps for boundaries}
        \State $V_{MSP} \gets \emptyset, E_{MSP} \gets \emptyset$
        \ForAll{$(p_i, p_j) \in \{(p_i, p_j) | p_i, p_j \in P, p_i \neq) p_j\}$}
            \State $V \gets \emptyset, E \gets \emptyset$
            \ForAll{$\tau \in T$}
            \Comment{Step 1}
                \State $V' \gets \Call{IntersectionPoints}{p_i, p_j, \tau}$
                \State \parbox[t]{\dimexpr\linewidth-\algorithmicindent}{$E' \gets \{(v_p, v_q) | (\tau(v_q) = \tau(v_p) + 1) \lor \\
                    (\tau(v_p) = \tau(v_q) = \min(\{\tau(v_r)|r \in V'\}))\}$ \strut}
                \State $V \gets V \cup V', E \gets E \cup E'$
            \EndFor

            \State $(V_{MSP}', E_{MSP}') \gets \Call{MinimumSpanningPath}{V, E}$
            \Statex
            \Comment{Step 2}
            \State $V_{MSP} \gets V_{MSP} \cup V_{MSP}'$
            \State $E_{MSP} \gets E_{MSP} \cup E_{MSP}'$
            
        \EndFor
        \State $D \gets \emptyset$
        \ForAll{$p \in P$}
        \Comment{Step 3}
            \State $E_p \gets \{e | e \in E_{MSP} \land p(e) = p\}$
            \State $d \gets \Call{SmallestEnclosingPolygon}{E_p}$
            \State $D \gets D \cup \{d\}$
        \EndFor
        \State{$D$ is the set of dominant regions for all players}
    \end{algorithmic}
\end{algorithm}

The first step is to compute the intersection points between the reachable regions of each pair of players. The intersection points at time $\tau_i$ are determined using a line segment intersection algorithm [5]. Each intersection point $v_j$ has a time-step attribute $\tau(v_j)$ denoting the time-step of the reachable region polygon that it was constructed from. In most cases, there will be zero or two intersection points between the polygons however degenerate cases exist where there are one, or three or more intersection points~\cite{horton-2013}. The intersection points $V = \{v_0, v_1, \ldots ,v_m\}$ between the pair of players for all time-steps are collected. A graph $G = (V, E)$ is constructed using $V$ as vertices and adding edges to $E$ between two points if either: the time-step of the two vertices are adjacent; or the two vertices have the same time-step and this time-step is the minimum of all intersection vertices:
\begin{multline*}
    E = \{(v_i, v_j)|(\tau(v_j) = \tau(v_{i})+1)~\lor \\
        (\tau(v_i) = \tau(v_j) = \min(\{\tau(v_k)|k \in V\}))\}
\end{multline*}

\begin{figure}
  \centering
    \includegraphics[width=0.7\linewidth, natwidth=300, natheight=277]{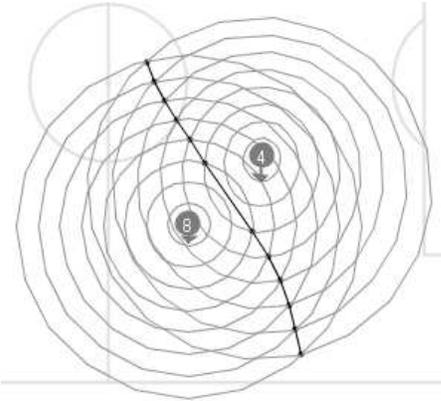}
  \caption{The intersection of time-step polygons defines the boundary between player's dominant regions.}
  \label{figure:ellipse-boundary-moving}
\end{figure}

The second step in the algorithm takes the graph $G$ as input and computes the reachable boundary between each pair of players. $G$ contains edges between each intersection point for consecutive time-steps. Typically each vertex in $V$ will have degree of four, with edges to two vertices whose time-step is immediately prior to the time-step of the current vertex, and two edges to the vertices whose time-step is immediately subsequent. The purpose of this step is to prune the graph so that each vertex in the graph has degree of no greater than two, and that the edges retained will be the shortest edges. The pruned graph will thus be a smooth path.
This is performed using a modified version of Kruskal's algorithm~\cite{kruskal1956}.
The algorithm is modified so that an edge $(v_i, v_j)$ is added to the output tree only if $\deg(v_i) < 2$ and $\deg(v_j) < 2$.  This modification means that the algorithm will return ``spanning paths'', i.e. the set of paths from the input graph that span the connected components of the graph. In most cases there will be a single path, see Figure~\ref{figure:ellipse-boundary-moving}. If there are two or more paths, select the one that contains an edge between two points $v_i$ and $v_j$ where $\tau(v_i) = \tau(v_j)$, as these are the intersection points with the minimum time-step and are thus closest to each player's site. The graphs for each pair of players are collected into a single graph $G_{MSP} = (V_{MSP}, E_{MSP})$. The edges in this graph are the line segments that form the reachable boundaries of each player. 

The final step of the algorithm is to compute the dominant region for each player by collecting the line segments that comprise the boundaries the player has with all other players, and then determining the polygon that encloses the site of the player. The algorithm shoots a ray from the site of the player and locates the closest intersecting line segment. 
The algorithm then collects the closest line segments by repeatedly ``turning left'' and walking to the next intersection with another line segment.
This loop stops when the original intersection point is again encountered. As such, this algorithm walks around the innermost polygon surrounding the player's site, and this polygon defines the player's dominant region. The dominant regions of all players induce a subdivision of the football pitch, as required, see Figure~\ref{figure:dominant-region}.

The following paragraph presents a sketch of the computational complexity of the approximation algorithm. The time and space complexity of the first two steps in the algorithm is $O(n \log{n})$ in the number of line segments in the reachable region boundaries. However, the complexity includes a large constant of $\binom{22}{2}$ as these steps must be carried out pairwise for all $22$ players. The worst-case complexity of the third step (determine the enclosing polygon) is quadratic on the number of line segments in the player motion model. For each line on the polygon boundary, we need to compute the closest intersecting line segment which is done by computing the intersection point for the line segment with every other line segment, which has linear complexity. The number of lines on the polygon boundary is at most $n$ thus yielding the quadratic complexity for this step, and for the overall algorithm. However, in practice this value is much smaller than $n$. In particular, the polygons of reachable regions will only intersect at zero or two points in most cases.

\section{Label Data}
\label{sec:data}

The supervised machine learning algorithms used in our experiments learn from training examples that have been labelled with the ground truth values. In this case, each pass event that occurs is labelled with a rating of its quality. This section describes the process used to capture and validate the label data.
 
\subsection{Labelling Process}

The labels used to train the classifiers were created by human observers watching video footage of the matches and rating the passes. Each observer individually watched video clips of the passes made and assigned a rating to each pass. The labels were chosen from a 6-point Likert Scale containing the following items: \textit{Very Good, Good, Slightly Good, Slightly Bad, Bad, Very Bad}.

The video clip for each pass included two seconds of footage preceding and following the pass event. The intention was to provide sufficient context to rate the pass, and also to ensure that each pass was rated in isolation, but not to include longer-term considerations such as the eventual outcome of a sequence of possession.


\subsection{Process Validation}

The classification process was designed using principles and techniques described by Lincoln and Guba~\cite{ls-ni-85}.

\begin{description}
  \item[Prolonged engagement:] Each observer viewed video clips of over 3,000 passes, consistent with Lincoln and Guba's technique of persistent observation. The intention was to ensure that the observers would identify characteristics that were most relevant to the quality and risk of a pass.

  \item[Member checking:] After the first two matches had been labelled, the difference between the labels assigned by the observers was determined. The passes where the distance between two observer's ratings was two or more were selected. The observers then viewed the footage of these passes together and discussed their reasoning. The purpose of this was to explore the characteristics of a pass that impacted the classifications, and was intended to ensure a consensus on the significant characteristics. 

  \item[Triangulation:] This describes the technique of using multiple sources of data to improve understanding. The labelling task was carried out by two observers so that the labels used in training the classifier were based on more than one source. The two labels for each pass were compared, and where there was not agreement, the label that was closest to the centre of the scale was selected as the ground truth. 
\end{description}

\subsection{Analysis of Classification Results}
\label{sec:data-analysis}

The ratings made by the observers are subjective - there is no \emph{correct} answer. It is thus natural to expect some disagreement between raters in their labels. Cohen's kappa~\cite{cohen-1968} is a heuristic measure of agreement between raters that takes into account the possibility that ratings can agree by chance, with a value of $0$ denoting chance agreement and $1$ denoting perfect agreement. Formally, Cohen's kappa is defined as:
\[
    \kappa = \frac{p_{o} - p_{e}}{1 - p_{e}}
\]

Here, $p_{o}$ is the fraction of observations where the raters were in agreement, and $p_{e}$ is the fraction of observations that could be expected to agree by chance.

The ratings made by the two observers yielded a Cohen's kappa of $0.393$ which indicates a moderate level of agreement, see Table~\ref{tab:cohens-kappa}. Although, Cohen's kappa does not have a statistical significance, it suggests that the observers produced similar outcomes. 

\section{Learning Algorithms}
\label{sec:classification}

The pass rating task is a classification problem and we evaluate several supervised machine learning algorithms for this. Informally, the goal of the learning algorithm is to produce a response function $h_\theta(\vec{x})$ that can predict the ground truth variable $y$ for a given input vector of predictor variables, $\vec{x}$. The algorithm is trained on the labelled example data with the objective of learning a parameterisation $\theta$ for the response function $h_\theta(\vec{x})$ such that the prediction error is minimised. 

The distribution of example data used in our experiments is unevenly distributed amongst the classes. The majority of examples were clustered towards the middle of the scale, see the frequency column of Table~\ref{tab:six-class-frequencies}. We thus selected learning algorithms designed to handle class imbalance. The experiments were run on three classifiers. In particular, we used multinomial logistic regression (MLR) with three different cost functions. We also examined a support vector machine (SVM) classifier and a RUSBoost classifier. The intention was to perform the experiments using diverse types of classifiers: SVM being a maximum margin classifier \cite{vapnik-1995} and RUSBoost is an ensemble method that utilises sampling and boosting of weak classifiers \cite{seiffert-2010}. 

\subsection{Multinomial Logistic Regression}

Multinomial logistic regression models the distribution of response variables using the multinomial distribution, conditioned on the input predictor variables. The probabilities are modelled using the softmax function.
\begin{align*}
    p(y=i|x;\theta) & = \frac{e^{\theta_i^{\top}x}}{\sum_{j=1}^{k}e^{\theta_i^{\top}x}}
\end{align*}
The parameterisation of $\theta$ is learned by minimising a regularised cost function $J(\theta)$, which is the sum of an empirical risk function $R(\theta)$ and a regularization term:
\begin{align}
    \label{eq:cost-function}
    \min_{\theta} J(X, y, \theta) & = \min_{\theta} R(X, y, \theta) + \lambda \lVert \theta \rVert _{p}
\end{align}
We evaluated three empirical risk functions. The function~\eqref{eq:risk-mle} is the risk function derived from the maximum likelihood estimation of $\theta$. The arithmetic~\eqref{eq:risk-arith} and quadratic~\eqref{eq:risk-quad} risk functions are intended to perform better under class imbalance conditions by computing the per-class risk \cite{liu-2011}. The arithmetic risk takes the sum of the per-class values, whereas the quadratic risk uses the root of the sum of the squared values.
\begin{align}
    \label{eq:risk-mle}
        R^{L}(X, y, \theta) & = - \frac{1}{m} \left[ 
            \sum_{i=1}^{m} \sum_{j=1}^{k} 1_{\{y^{(i)} = j\}} 
            \log \frac{e^{\theta_j^{\top}x^{(i)}}}{\sum_{l=1}^{k}e^{\theta_l^{\top}x^{(i)}}} 
          \right] \\
        \label{eq:risk-arith}
        R^{A}(X, y, \theta) & = - \frac{1}{k} \sum_{j=1}^{k} \left[ 
                \frac{1}{m_j} \sum_{i=1}^{m_j} 1_{\{y^{(i)} = j\}} 
                \log \frac{e^{\theta_j^{\top}x^{(i)}}}{\sum_{l=1}^{k}e^{\theta_l^{\top}x^{(i)}}}
            \right] \\
        \label{eq:risk-quad}
        R^{Q}(X, y, \theta) & = -\sqrt{\frac{1}{k} \sum_{j=1}^{k} \left[ 
                \frac{1}{m_j} \sum_{i=1}^{m_j} 1_{\{y^{(i)} = j\}}
                \log \frac{e^{\theta_j^{\top}x^{(i)}}}{\sum_{l=1}^{k}e^{\theta_l^{\top}x^{(i)}}}
            \right]^{2}}
\end{align}
The regularization term includes the norm of $\theta$. We evaluate the $\ell_1$- and $\ell_2$-norms in our experiments, i.e. $p \in \{1, 2\}$ in Equation~\ref{eq:cost-function}. Moreover, the $\ell_1$-norm will induce a sparse parameterisation of $\theta$~\cite{jenatton-2011}, and we investigate the predictor variables whose corresponding value in $\theta$ is non-zero as a measure of the importance of the predictor variable.

\section{Experiments}
\label{sec:experiments}

We carried out a set of experiments to answer the following questions:
\begin{enumerate}
    \item{Is it possible to find a classification function and set of predictor variables to accurately predict the quality of a pass?}
    \item{Do the predictor variables that are computed using complex algorithms from computational geometry necessarily lead to better prediction?}
\end{enumerate}

\subsection{Setup}

The objective of the experiments conducted for this research was to learn an optimal set of parameters $\theta^{\ast}$ that characterize a response function $h_{\theta}(x)$ such that the response function makes correct predictions on unseen examples. In order to compare the relative performance of the learned response functions, we held out 20\% of the labelled data for testing. The held-out test examples were stratified so that the per-class frequency of examples in the test set was consistent with the distribution in the training examples. The classifiers were trained using tenfold cross-validation of the training examples, and evaluated comparing the classifier prediction to the ground-truth label on the test examples.

The experiments were conducted using two labelling schemes. The six-class scheme used the labels that were assigned by the observers, see Section~\ref{sec:data}. The three-class scheme aggregated the six classes into three classes: \emph{Good}, \emph{OK} and \emph{Bad}. Tables~\ref{tab:six-class-frequencies} and~\ref{tab:three-class-frequencies} summarise the distribution of examples to classes.

There are several methods of evaluating the performance of the response function, and we evaluated the response functions using the following objective functions. 
\vspace{-2mm}
\begin{itemize}
    \item{Accuracy} 
        \vspace{-2mm}
    \item{Precision}
        \vspace{-2mm}
    \item{Recall}
        \vspace{-2mm}
    \item{$F_{\beta=1}$-Score}
\end{itemize}

The metrics for precision, recall and $F_{\beta=1}$-Score are necessarily calculated on a per-class basis, so a simple mean of the per-class values was used as the objective function in these cases.


\begin{table}
    \centering
    \small
    \caption{Per-class frequencies for six-class model}
    \label{tab:six-class-frequencies}
        \begin{tabular}[t]{l c r}
            \hline
            Class         & Relative Frequency & Count \\
            \hline
            Very Good     & 0.008 &   24 \\
            Good          & 0.058 &  171 \\
            Slightly Good & 0.624 & 1829 \\
            Slightly Bad  & 0.206 &  604 \\
            Bad           & 0.088 &  257 \\
            Very Bad      & 0.016 &   47 \\
            \hline
        \end{tabular} 
        \hspace{2em}
\end{table}
\begin{table}
    \centering
    \small
    \caption{Per-class frequencies for three-class model}
    \label{tab:three-class-frequencies}
        \begin{tabular}[t]{l c r}
            \hline
            Class   & Relative Frequency & Count \\
            \hline
            Good   & 0.066	 &   193 \\ 
            OK     & 0.789	 &  2314 \\ 
            Bad    & 0.145	 &   425 \\ 
            \hline
        \end{tabular}
\end{table}

\subsection{Results}

The results of the experiments are summarised in Table~\ref{tab:6-class-results} and \ref{tab:3-class-results}. 
The overall accuracy of the classifiers shows that SVM and MLR using the simple cost function are the optimal configurations. The multinomial logistic regression classifiers using the arithmetic and quadratic loss functions showed improved recall, but this was at a cost of precision. This is in line with expectations, as these cost functions are less sensitive to the class imbalance, and would be more likely to assign examples to the minority classes. However, the improvement in recall was not sufficient to outweigh the reduction in precision, and so the overall accuracy was also reduced.

\begin{table}
    \caption{Summary results for 6-class model}
    \label{tab:6-class-results}
    \centering
    \small
    \begin{tabular}{l c c c c c}
        \hline
        Classifier& Accuracy& Precision& Recall & $F_{\beta=1}$-Score \\
        \hline
        MLR       & 0.669	& 0.462	& 0.417	& 0.455	\\
        MLR-Arith & 0.520	& 0.337	& 0.466	& 0.401	\\
        MLR-Quad  & 0.527	& 0.345	& 0.429	& 0.408	\\
        SVM       & 0.660	& 0.395	& 0.327	& 0.358	\\
        RUSBoost  & 0.580	& 0.359	& 0.453	& 0.348	\\
        \hline
    \end{tabular}
\end{table}

\begin{table}
    \caption{Summary results for 3-class model}
    \label{tab:3-class-results}
    \centering
    \small
    \begin{tabular}{l c c c c}
        \hline
        Classifier& Accuracy& Precision& Recall & $F_{\beta=1}$-Score \\
        \hline
        MLR       & 0.829	& 0.666	& 0.752	& 0.638	\\
        MLR-Arith & 0.730	& 0.580	& 0.780	& 0.612	\\
        MLR-Quad  & 0.741	& 0.581	& 0.784	& 0.617	\\
        SVM       & 0.858	& 0.713	& 0.734	& 0.711	\\
        RUSBoost  & 0.756	& 0.600	& 0.781	& 0.646	\\
        \hline
    \end{tabular}
\end{table}

\section{Analysis}
\label{sec:analysis}

The experimental results show that we are able to learn a classifier to perform the pass classification task. We provide an analysis of the results in this section.

\subsection{Classifier Performance}

The experimental results in both the three-class and six-class settings show that it is possible to learn a classifier that performs better than random, and also better than simply selecting the class with the most examples assigned to it, the so-called majority classifier. The error present in these classifiers appears to be the result of bias, as the accuracy of the classifiers on the unseen test data is close to the accuracy on the training data. For example, the SVM classifier in the six-class setting has an accuracy of $66.0\%$ on the test data and $66.6\%$ on the training data. This suggests that the predictor variables used in the classifier do not contain sufficient information to accurately make the predictions, see Section~\ref{sec:sources-of-error}.

\subsection{Predictor Variable Importance}

The predictor variables were computed by feature functions of varying complexity. This begs the question whether the work involved in the complex feature functions resulted in improved performance. The task of determining the importance of a particular feature is not obvious, however we investigated a number of approaches to this. 

It should be noted that none of the predictor variables used in our experiments was a strong predictor of the quality label assigned to the pass. A visual examination of scatter-plots of pairs of response variables did not reveal any pairs that clearly separated the labelled data. Moreover, applying principal component analysis produced a coefficient of $0.03$ on the first eigenvector. Therefore, the importance of any given feature is likely to be incremental.

We selected the features that were computed using the dominant region described in Section~\ref{subsection:dominant-region}. Six of the 49 features constructed were based on the dominant region. We investigated the importance that various algorithms assigned to these six features.

The first approach was to investigate the weights learned when using $\ell_1$ regularization, as this approach is sparsity-inducing \cite{jenatton-2011}. Informally, this means that features that are of low importance will have a zero weight and important features will be non-zero. We examined the weights vector for the MLR classifier using $\ell_1$ regularization. In the three class setting, seven variables were assigned non-zero weights and three of these were based on the dominant region. However, in the six-class setting, only one out of eleven variables were assigned non-zero weights.

Furthermore, we considered the predictor importances derived from the RUSBoost classifier. In this case, the predictor variable of the dominant region of the passer was ranked the most important variable in the three- and six-class setting. However, the next most important variable based on the dominant region was seventeenth.

While some of the dominant region predictor variables are important to the classifiers, their importance to the classifiers is not clear. Given the overall bias in the model, there is clearly scope to improve the input predictor variables, and this is something that we plan to continue researching.

\subsection{Inter-Rater Agreement}
\label{sec:rater-agreement}

In Section~\ref{sec:data-analysis} we discussed the fact that it is reasonable to expect that the observers would rate passes differently, and we applied Cohen's kappa \cite{cohen-1968} as a heuristic to evaluate the inter-rater agreement between the observers. Here we extend that analysis to the responses produced by the classifiers in the experiments. We computed Cohen's kappa in a pairwise manner on all labellings, see Table~\ref{tab:cohens-kappa}. The objective was to examine whether the inter-rater agreement between an observer and a classifier was significantly different from that between two observers. 

Comparing the observer-classifier kappa's, there is a significant difference between the values for the two observers, where Observer~1 has higher agreement with the classifiers. In particular, the value of Cohen's kappa is greater between Observer~1 and the SVM classifier than between Observer~1 and Observer~2. The kappa values for Observer~1 with the other classifiers were of similar magnitude to the kappa between the two observers. This interestingly suggests that the classifier will produce results that are as similar to an observer's results as the similarity between the results of two observers. 

\begin{table*}
    \caption{Cohen's Kappa showing inter-rater agreement between human observers and learned classifiers in six-class setting.}
    \label{tab:cohens-kappa}
    \small
    \centering
    \begin{tabular}{l c c c c c c c}
        \hline
        ~           &Observer~1&Observer~2& MLR      & MLR-Arith& MLR-Quad & SVM      & RUSBoost\\
        \hline
        Observer~1  &   -  	& 0.393	& 0.360	& 0.323	& 0.325	& 0.464	& 0.338	\\
        Observer~2  & 0.393	&   -  	& 0.232	& 0.205	& 0.214	& 0.309	& 0.222	\\
        MLR         & 0.360	& 0.232	&   -  	& 0.410	& 0.424	& 0.598	& 0.350	\\
        MLR-Arith   & 0.323	& 0.205	& 0.410	&   -  	& 0.894	& 0.343	& 0.450	\\
        MLR-Quad    & 0.325	& 0.214	& 0.424	& 0.894	&   -  	& 0.351	& 0.464	\\
        SVM         & 0.464	& 0.309	& 0.598	& 0.343	& 0.351	&   -  	& 0.366	\\
        RUSBoost    & 0.338	& 0.222	& 0.350	& 0.450	& 0.464	& 0.366	&   -  	\\
        \hline
    \end{tabular}
\end{table*}

\subsection{Sources of Error}
\label{sec:sources-of-error}

The purpose of the feature functions described in Section~\ref{sec:feature-engineering} is to capture sufficient information to train a classifier that has good predictive performance. The classifiers described in this paper tend to exhibit high bias, and thus it is worthwhile to examine the sources of error in the model.
\begin{description}
  \item[Sample Bias]
    The trajectory data available is limited to four matches. All of these matches are home matches for Arsenal Football Club. Arsenal were a strong team in the 2007/08 season, finishing third. They were unbeaten at home, playing $19$ games, winning $14$ and drawing $5$. The opposition teams in the four matches were Aston Villa (finished $6$-th in season), Blackburn Rovers ($7$-th), Bolton Wanderers ($16$-th) and Reading ($18$-th). Given that teams will often vary their tactics based on whether they are playing home or away, and also in terms of the relative strength of the opposition, then there is the possibility of bias here.
\item[Source Data] 
    The data used to produce the predictor variables is limited to trajectory data for the players and event sequence. The learning algorithm must train the classifier using only this information. However, the observers who labelled the training data may consider an number of other aspects when making their rating, for example the aesthetics of the pass, their prior belief about the player making the pass, or the apparent intensity of the current state of the match.
  \item[Video Framing] The labelling made by the observers was performed by viewing video footage of the match. This footage does not display the entire playing field, and thus the observer cannot take into account the state of players not in the video frame. The classifier, however, does not know what parts of the playing field were visible to the observer, and thus cannot discriminate based on this.
  \item[Facing direction of players]
      The source data used to construct the predictor variables only provides the location of each player. The orientation of the player in not available, and thus must be extrapolated. There are two plausible extrapolations that were used: that the player is facing in the direction of motion; or that the player is facing the ball.
  \item[Ball trajectory]
      The trajectory of the ball is not provided, and thus is extrapolated by using the event data to determine when a player touches the ball. In between such events, the location and speed of the ball is interpolated using a simple linear model. This is clearly an approximation, as the ball may not travel in a straight line, for example if it is kicked in the air. Moreover, the velocity of the ball will not remain constant between events as is the case in the model.
  \item[Approximation Algorithm] The approximation algorithm used to construct the dominant region necessarily introduces an error in the predictor variables that are computed using the algorithm.
\end{description}

The training error that was inherent in all classifiers points to a deficiency of information in the constructed features and these apparent sources of error reinforce this view.

\section{Conclusion}
\label{sec:conclusion}

In this paper we present a model that is able to learn a classifier to rate the quality of passes made during a football match with an accuracy of up to $85.8\%$. We compared the ratings made by the classifier with those made by human observers, and found that the level of agreement between the machine classifier and an observer was similar in magnitude to the level of agreement between two observers.

The model uses feature functions based on methods from computational geometry. We present an efficient approximation algorithm for computing the dominant region. This structure is intended to provide information about the strategic and physiological state of the match, however it is costly to compute. We evaluate the importance to the classifier of the predictor variables based on the dominant region, and find them to be of moderate importance.

The model and experiments described provide a foundation for further research, and several areas to investigate became apparent during this research, in particular, the design of new predictor variables to mitigate the sources of error inherent in the model. These are issues that we plan to explore further in subsequent research.

\bibliographystyle{abbrv}
\bibliography{pass_classification_paper}  
\balancecolumns
\end{document}